# HALF OF AN IMAGE IS ENOUGH FOR QUALITY ASSESSMENT

*Junyong You[1], Yuan Lin[2], Jari Korhonen[3]*

1. NORCE Norwegian Research Centre, Bergen, Norway; 2. Kristiania University College, Bergen, Norway; 3. University of Aberdeen, Aberdeen, UK

## ABSTRACT

Deep networks have demonstrated promising results in the field of Image Quality Assessment (IQA). However, there has been limited research on understanding how deep models in IQA work. This study introduces a novel positional masked transformer for IQA and provides insights into the contribution of different regions of an image towards its overall quality. Results indicate that half of an image may play a trivial role in determining image quality, while the other half is critical. This observation is extended to several other CNN-based IQA models, revealing that half of the image regions can significantly impact the overall image quality. To further enhance our understanding, three semantic measures (saliency, frequency, and objectness) were derived and found to have high correlation with the importance of image regions in IQA.

## 1. INTRODUCTION

Image Quality Assessment (IQA) plays a crucial role in determining the perceived quality of images with distortions, thereby ensuring the best Quality of Experience for end users. With the massive growth in User Generated Content (UGC) produced by mobile devices, the demand for No-Reference IQA (NR-IQA) models has skyrocketed [1].

Traditional NR-IQA models have focused on deriving features that represent individual distortion types [2][3]. However, UGC images often contain diverse distortions, making it challenging to derive specific features for each type. A recent approach [4] suggests that distortions can cause images to look unnatural, leading to the development of IQA models based on natural scene statistics [5][6].

With the rise of deep learning, Convolutional Neural Networks (CNNs) have become the dominant tool in image analysis tasks [7][8]. Consequently, they have also gained widespread interest in the IQA research community. Early works attempted to adapt popular image recognition models, such as VGGNet and ResNet, for quality prediction [9]-[11]. For instance, Gao *et al.* [9] used a pretrained VGGNet to extract image features, which were then used as input for a Support Vector Regression model to predict quality. Hosu *et al.* [11] used InceptionResnetV2 followed by Fully Connected layers for regression, achieving promising results on a large-scale IQA dataset they produced. Furthermore, CNNs can also be used to derive features at multiple scales using multi-stage architectures, providing abundant information for IQA [12][13].

As a perceptive measure, image quality is affected by visual mechanisms, e.g., attention and contrast sensitivity [14]. Thus, IQA models simulating visual mechanisms have also been proposed. A typical example is to include attention mechanism in quality prediction, that can be performed by CNNs [15] or analytic approach [16]. Due to its inherent capability to represent a self-attention mechanism, transformer has also been applied in IQA [17][18]. As an early effort, we have proposed a TRIQ model by using a transformer encoder on features derived from CNN backbone [17], which can benefit from the inductive bias of CNNs to overcome the problem that IQA datasets are relatively small-scale.

Despite the impressive performance demonstrated by CNNs, transformers, and their hybrid models in IQA, few studies have attempted to explain how these models work. In the image classification domain, explainable AI (XAI) has primarily focused on generating local heat maps using methods like LIME [19][20]. The visualization of class activation maps using gradient-based localization has also been used to explain CNNs [21]. Additionally, the self-attention nature of the transformer architecture makes it easier to visualize attention maps, as seen in the ViT model [22]. In IQA models, there have only been a few studies [23][24] that focus on specific scenarios, such as medical treatment, and use methods like Grad-CAM for explanation.

We have tested the LIME and Grad-CAM methods on several IQA models, including TRIQ [17], KonCept [11], and the positional masked TRIQ that will be presented in the next Section. LIME generates salient regions from the model predictions. In addition, as TRIQ uses a hybrid model consisting of CNN and transformer, we multiplied the heatmap generated by Grad-CAM from the CNN backbone with the attention weights from the transformer to derive an attention map illustrating the importance degrees of image regions for IQA. Fig. 1 shows two example results demonstrating that the two methods do not explain the TRIQ model properly. LIME intends to identify local and discrete regions, and Grad-CAM often produces attention regions irrelevant to IQA.

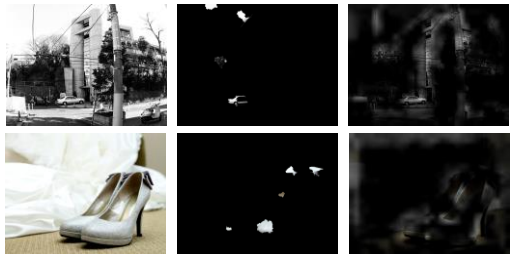

**Fig.1. Images (left), LIME map (middle) and Grad-CAM map (right) derived from the TRIQ model.**

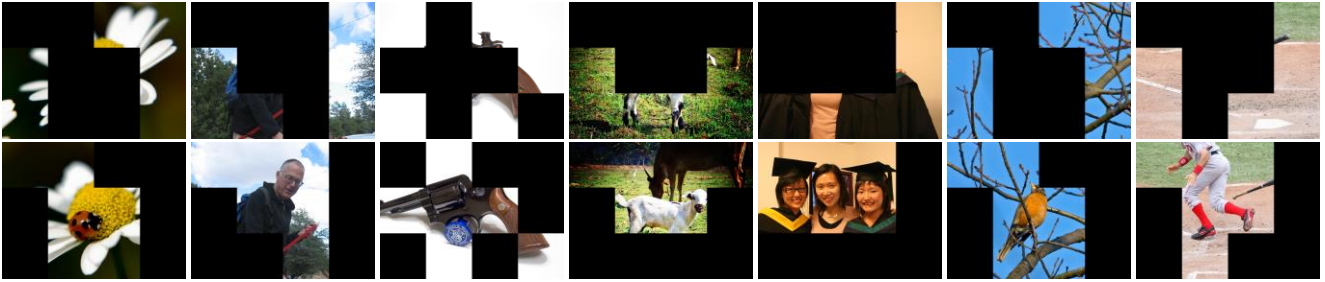

**Fig. 2.** Example images with important and trivial regions masked. Trivial regions (top row) indicate these regions (non-black regions) are trivial for IQA performed by TRIQ, while important regions (bottom row) are the opposite.

In this work, we first investigate the importance levels of individual image regions on image quality by a positional masked transformer model, as presented in Section 2. Our observations reveal that almost half of an image might contribute trivially to IQA, whereas the other half is crucial. By comparing the important regions against the trivial regions, a connection between quality perception and image semantics explained in Section 3 has been identified. The experiments reported in Section 4 confirm the above descriptions.

## 2. POSITIONAL MASKED TRIQ FOR IQA

In our earlier work, a transformer for IQA (TRIQ) has been proposed, following a similar approach to ViT, where a hybrid model consisting of a CNN backbone and a transformer encoder is employed [17]. The CNN without the head layer produces a feature matrix that is fed into the transformer encoder for quality prediction. CNN derives image features in a local manner. For example, ResNet50 without the head layer generates a feature matrix of size [$H/32$, $W/32$, 2048] on an image with resolution of [$H$, $W$, 3]. In the feature matrix (*FM*), a vector with size of [1, 1, 2048] at each location roughly represents certain local region in the original image. Therefore, it is possible to investigate the impact of local regions on the overall predicted image quality by manipulating the feather vectors in the feature matrix.

We insert a masking operation in the transformer encoder. The masking operation is performed by adding a very high negative scalar value (-1e9) to the dot product of query and key in the self-attention block at the masked locations. Consequently, the Softmax function produces output close to zero at the masked locations. In other words, the masked locations make almost no contributions to the final output. Therefore, with this approach, we can find out which feather vectors in *FM* corresponding to individual regions in the original image have significant or trivial impact on image quality.

Using the KonIQ-10k dataset [11] as an example, all the images have a resolution of 768×1024. Consequently, the size of the feature matrix *FM* is [24, 32, 2048]. We divide the feature matrix into a 3×4 grid, where each block contains 8×8 =64 feature vectors, and each vector has 2048 features. We manually mask *n* blocks, where *n*=1, 2, …, 11, respectively. Masking one block means that the 64 feature vectors in this block do not contribute to the prediction of image quality. On the other hand, masking 11 blocks means that only 64 feature vectors in that unmasked block contribute to quality prediction. Note that *n* loops through all the possible combinations of $C_{12}^n$. The image quality values are computed by the masked TRIQ model, which are then compared with image quality predicted by TRIQ without the masking operation.

The comparison in terms of Pearson correlation (PLCC) can indicate impact of those *n* blocks on the model prediction. A lower PLCC means that the masked *n* blocks are more important for IQA performed by the positional masked TRIQ model because the prediction has been significantly changed. Finally, an average of all the PLCC values from the grid combinations is computed and used as an importance indicator of each block on the masked TRIQ prediction. The 12 blocks are divided into two groups evenly: important and trivial, based on the PLCC values. The important group means that the contained 6 blocks can significantly change the TRIQ prediction if they were masked, whereas the trivial one is the opposite. We have conducted an experiment on all the 10,073 images in the KonIQ-10k dataset and obtained a correlation of 0.97 between the TRIQ predictions without and with masking the trivial blocks, and a correlation of 0.67 when masking the other half of the blocks. A close result, i.e., 0.96 vs. 0.68, was also obtained on the SPAQ dataset [25].

As explained above, CNN extracts local features. Thus, a block corresponds to a local image region, and a half of the 12 blocks corresponds to half of an image approximately. Consequently, the above observation implies that half of an image contributes trivially to IQA by the masked TRIQ model, compared to the other half. Fig. 2 shows several image examples with masking the important and trivial halves, and ("PLCC on test sets" indicate the performance of individual models, "Important regions" means zeroing the important regions, "Trivial regions" zeroing the trivial regions, "Predicted" is the PLCC between model predicted quality values on the original images and the zeroed images on the test sets; and "MOS" indicates the PLCC between the model predicted quality on zeroed images and the MOS values on original images on the test sets) interested readers can refer to [26] for all the images with important and trivial regions masked.

**Table I. Impact evaluation of important and trivial regions on CNN-based IQA models**

| Models | KonIQ-10k | | | | | SPAQ | | | | |
| --- | --- | --- | --- | --- | --- | --- | --- | --- | --- | --- |
| | PLCC on test sets | Important regions | | Trivial regions | | PLCC on test sets | Important regions | | Trivial regions | |
| | | Predicted | MOS | Predicted | MOS | | Predicted | MOS | Predicted | MOS |
| TRIQ | 0.922 | 0.680 | 0.671 | 0.972 | 0.902 | 0.916 | 0.695 | 0.675 | 0.969 | 0.907 |
| AIHIQnet | 0.929 | 0.816 | 0.804 | 0.959 | 0.896 | 0.928 | 0.810 | 0.801 | 0.958 | 0.912 |
| Koncept | 0.916 | 0.817 | 0.813 | 0.960 | 0.901 | 0.831 | 0.802 | 0.788 | 0.957 | 0.811 |
| DBCNN | 0.856 | 0.804 | 0.796 | 0.963 | 0.887 | 0.894 | 0.808 | 0.797 | 0.960 | 0.868 |
| Swin-IQA | 0.956 | 0.835 | 0.811 | 0.974 | 0.921 | 0.933 | 0.804 | 0.782 | 0.963 | 0.902 |

The conclusion above is drawn for the masked TRIQ model, inspiring us to investigate if it is valid for other CNN-based IQA models as well. In other words, is half of an image always enough for a CNN-based IQA model? Our experiments reported in Section 4 aim at answering this question.

## 3. SEMANTIC IMAGE MEASURES

The positional masked TRIQ can identify the important and trivial regions for IQA. However, a remaining problem is to explain why a region is important or not for IQA. This question cannot be answered by analyzing the IQA model itself. Therefore, we have carefully investigated all the 10,073 KonIQ-10k dataset images with the important and trivial regions identified. Based on the investigation, we observed that there is a significant difference in three semantic measures between the important and trivial regions.

a) *Saliency* describes the degrees of regions in an image where viewers fixate with high priority. We have observed that the regions important for IQA often contain salient objects, e.g., human faces, heads of animals, or birds on a tree.

b) *Spatial frequency* is a characteristic of patterns that are periodic across position in space. Such characteristic is heavily related to contrast sensitivity mechanism. We observed that the important regions often contain objects with mid-range of spatial frequencies. In other words, the plain regions (i.e., low frequency) and complex texture regions (i.e., high frequency) are rarely identified as important regions for IQA. This observation is in accordance with the contrast sensitivity function (CSF) that the viewing sensitivity increases with the spatial frequency, peaks at certain point and then gradually decreases with increasing the frequency [27].

c) *Objectness* measures the likelihood of a group of pixels in an image to be an object, especially a foreground object. A higher objectness level indicates that the pixel group more probably belongs to an object. We have observed that the most important regions contain meaningful objects, rather than pure background.

The above three measures are understandable, indicating easy explainability that TRIQ or other CNN-based IQA models cannot provide. However, to quantitatively demonstrate if the three measures are in accordance with quality perception, quantitative representations of the measures are requisite.

Saliency measure can be derived from the spectral residual of fast Fourier transform (FFT) that prompts the regions with unique appearance in an image [28]. The saliency map $S$ of an image is roughly explained as follows:

$$S = G \cdot F^{-1}[\exp(R + P)]^2 \quad (1)$$

where $F$ denotes FFT of the image, $R$ and $P$ are the spectral residual and the phase spectrum of the image, respectively, and $G$ is a Gaussian filter for smoothing the output [28].

The spatial frequency measure is also derived from FFT. The maximal spatial frequency (cycles/degree) is set to 50 that is around the highest frequency that the human visual system (HVS) can perceive according to the contrast sensitivity function derived from visual experiments [27][29]. Subsequently, the frequency range [0, 50] is divided into 10 bins, and the magnitudes of frequency responses from the FFT result at each pixel are summed in each segment, which is used as the spatial frequency measure of the pixel. Finally, the objectness measure is obtained using the appearance stream by a fully convolutional network in [30], generating a pixel objectness map for an image.

## 4. EXPERIMENTS

The experiments include two parts: validation on whether half of an image is sufficient for CNN-based IQA models, and if the proposed semantic measures can appropriately detect important and trivial regions in an image for IQA.

### 4.1. Is half of an image enough for IQA?

Four CNN-based IQA models were used in this experiment, including TRIQ [17], AIHIQnet based on multi-scale structure of CNN [31], KonCept [11], and DBCNN [32].

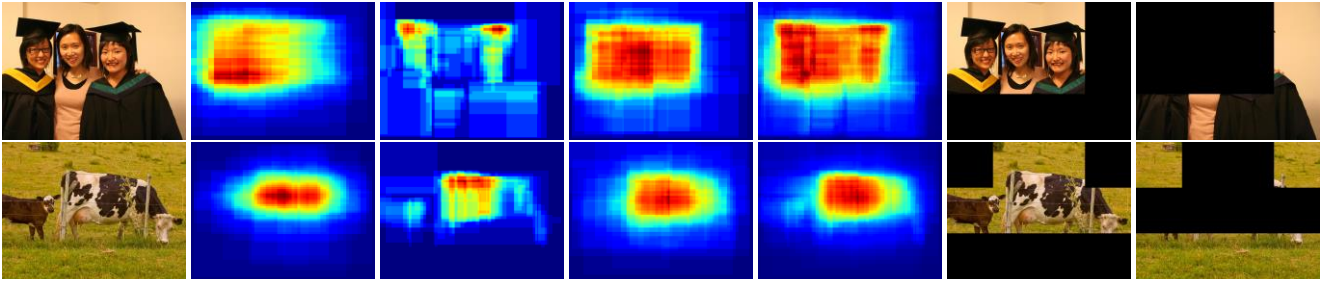

**Fig. 3.** Semantic measures of images, and important and trivial regions, from left to right: image, saliency measure, spatial frequency measure, objectness measure, averaged measure, important regions, and trivial regions.

**Table II.** Matching degrees (in percentage) between semantic measures and region importance for IQA at different thresholds ($T$) on the combined KonIQ-10k and SPAQ dataset

| Measures | $T$=1 | $T$=2 | $T$=3 | $T$=4 | $T$=5 |
|---|---|---|---|---|---|
| Saliency | 100 | 96 | 69.2 | 52.4 | 37.8 |
| Frequency | 100 | 99 | 70.1 | 60.1 | 40.8 |
| Objectness | 100 | 100 | 77.3 | 66.6 | 44.7 |
| Averaged | 100 | 100 | 79.5 | 65.8 | 45.6 |

Swin-IQA [33] using Swin [34] for IQA has also been tested. The models were first trained on two IQA datasets: KonIQ-10k and SPAQ. The datasets were split into train, validation and test sets, respectively. To avoid the long tail problem, we carefully performed the splits such that the image quality and content complexity levels were evenly distributed across train/ validation/test sets. Commonly used training tricks were employed, e.g., learning rate scheduler, image augmentation using horizontal flip only as other methods can potentially change the perceived image quality. PLCC between the predicted image quality and ground-truth on the validation set was monitored to obtain the best weights that will be used in the experiment for each model.

The approach explained in Section 2 was used to identify half of an image as important and the other half as trivial for IQA. Subsequently, we have manually set the pixel values to zero in important and trivial regions, respectively. Note that the pixels are zeroed after an image has been preprocessed via e.g., normalization. Due to local operation of CNN, such assignment will be transferred to the chain of convolution computations in a deep CNN, and in turn it will affect the followed computations (e.g., transformer encoder), even though it does not equal to directly masking out half of the input to the followed computations.

Table I reports the results in terms of PLCC values between quality predicted by the IQA models on the original images and the zero-assigned images. The results clearly demonstrate that assigning zeros to the pixels in the important regions significantly affects quality prediction by these IQA models, with PLCC≈0.80. On the other hand, zeroing the trivial regions only marginally affects the IQA models with high PLCC≈0.96. These models can still provide sufficiently accurate quality predictions even though half images (i.e., the trivial regions) are omitted, e.g., see the PLCC values between the model prediction and the MOS values when the trivial regions are zeroed. This validates our observation that half of an image is enough for CNN-based IQA models, at least concerning the models tested in our experiment.

### 4.2. Are the semantic measures matching IQA?

Three semantic measures are proposed in Section 3, and we also use a fourth measure by normalizing them into the same range and then averaging. Fig. 3 illustrates two images and the respective heatmaps of the measures. In addition, the important and trivial regions are also shown in Fig. 3, showing high accordance between the semantic measures and the region importance.

We also performed a quantitative analysis on all the images in KonIQ-10k and SPAQ datasets. Each image was divided into 12 (3×4) regions evenly, and then the 6 important regions and 6 trivial regions were identified by the approach explained in Section 2, respectively. Subsequently, the averages for the four pixel-wise semantic measures were computed for each of the 12 regions. Then, we counted the number of regions correctly classified as important or trivial by the semantic measures. According to a threshold $T$ (=1, 2, 3, 4, 5), an image is counted as matched by a measure if $n$ regions ($n>T$) are identified as important regions and they also belong to the 6 regions with higher semantic measures, meanwhile the other $n$ regions ($n>T$) are identified as trivial regions and in the other 6 regions with lower semantic measures. Such approach can evaluate if a region identified by a semantic measure is also important or trivial for IQA. Table II reports the matching degrees of the semantic measures at different thresholds $T$. Taking an example of $T$=4, the result shows that for 66.6% images in the two datasets, at least 5 regions wither higher values of objectness measure belong to the 6 important regions for IQA, whilst other 5 or 6 regions with lower objectness levels are trivial for IQA. The matching degrees also confirm that the employed semantic measures indeed reveal relevant image attributes for quality perception.

Finally, another experiment was conducted using the averaged measure to identify important and trivial regions and then zeroing either important or trivial regions. The five IQA models in Section 4.1 have been run again to calculate the correlation between the original prediction and the prediction on zeroed images. An average PLCC of 0.803 when zeroing the important regions, and PLCC of 0.885 when zeroing the trivial regions was obtained over the IQA models. This also confirms that the semantic measures can identify the important and trivial regions for IQA at a high accuracy.

## 5. CONCLUSION

An approach to identify important and trivial regions for CNN-based IQA models has been proposed, based on a positional masked TRIQ (transformer for image quality) model. This approach validates that half of an image seems to be sufficient for an IQA model using CNN as backbone. Furthermore, three semantic measures, namely saliency, spatial frequency, and objectness, were derived, showing to be accordant with the important or trivial regions for IQA with a high accuracy. This approach is helpful to analyze how IQA models make their prediction, which can further improve explainability of deep learning models for IQA.